\documentclass{article}
\usepackage{spconf,amsmath,graphicx}
\usepackage{amsfonts}
\usepackage{booktabs,multirow}   
\usepackage{hyperref}

\usepackage{microtype}
\title{Investigating Identity Signals in Conversational Facial Dynamics \\via Disentangled Expression Features}

\name{Masoumeh Chapariniya$^{1,2}$, 
      Pierre Vuillecard$^{3,4}$,  
      Jean\mbox{-}Marc Odobez$^{3,4}$, 
      Volker Dellwo$^{1}$,
      Teodora Vukovi\'{c}$^{1,2}$,}

\address{$^{1}$Department of Computational Linguistics, University of Zurich , Zurich, Switzerland\\
$^{2}$UZH Digital Society Initiative\\
$^{3}$Idiap Research Institute, Martigny, Switzerland\\
$^{4}${\'E}cole Polytechnique F{\'e}d{\'e}rale de Lausanne (EPFL), Lausanne, Switzerland}

\begin{document}
\ninept  
\maketitle

\begin{abstract}
This work investigates whether individuals can be identified solely through the pure dynamical components of their facial expressions, independent of static facial appearance. 
We leverage the FLAME 3D morphable model to achieve explicit disentanglement between facial shape and expression dynamics, extracting frame-by-frame parameters from conversational videos while retaining only expression and jaw coefficients. 
On the CANDOR dataset of 1,429 speakers in naturalistic conversations, our Conformer model with supervised contrastive learning achievese 61.14\% accuracy on 1,429-way classification—458 times above chance demonstrating that facial dynamics carry strong identity signatures. We introduce a drift-to-noise ratio (DNR) that quantifies the reliability of shape expression separation by measuring across-session shape changes relative to within-session variability. DNR strongly negatively correlates with recognition performance, confirming that unstable shape estimation compromises dynamic identification. Our findings reveal person-specific signatures in conversational facial dynamics, with implications for social perception and clinical assessment.
\end{abstract}
\begin{keywords}
Facial expression dynamics, person identification, FLAME model, Supervised Contrastive Learning, Conformer
\end{keywords}
\section{Introduction}

Human facial expressions during natural conversation contain remarkable individual specificity~\cite{dobs2016identity}. Each person exhibits unique patterns in how they smile, speak, and emote—behavioral signatures that persist across contexts and interactions. Recent psychological research demonstrates these individual differences fundamentally shape social perception through distinct ``expression perceptive fields'' that determine how people categorize and interpret emotions~\cite{murray2024expression}.

This individuality in facial dynamics has profound implications across multiple domains. Clinical assessments struggle to separate person-specific expression patterns from neurological symptoms~\cite{yoonesi2024facial}, potentially leading to misdiagnosis. Social interaction studies cannot disentangle individual style from emotional content~\cite{barrett2019emotional}, limiting our understanding of affective communication. Technology systems fail to capture personalized expression dynamics~\cite{thambiraja2023imitator}, resulting in uncanny or generic avatar animations.

The fundamental challenge lies in methodology: how can we rigorously isolate dynamic facial patterns from static facial structure? Prior computational approaches using keypoint sequences~\cite{papadopoulos2022facegcn} or spatiotemporal networks~\cite{kay2023person} inherently conflate appearance with motion, making it impossible to determine whether models learn true behavioral signatures or merely view-invariant representations of facial geometry. This entanglement particularly impacts applications requiring personalized understanding—from distinguishing clinical markers in autism and Parkinson's disease~\cite{drimalla2021imitation,bandini2021new} to creating authentic avatars~\cite{thambiraja2023imitator}.

We address this challenge through explicit disentanglement using the FLAME parametric model~\cite{li2017flame}, which provides mathematically separated representations of shape and expression. By extracting FLAME parameters frame-by-frame and retaining only expression and jaw coefficients while completely discarding static shape, we ensure our analysis focuses solely on dynamic behavior. These pure dynamics sequences are processed through various temporal architectures, with particular focus on Conformer models that combine self-attention with convolution for multi-scale temporal modeling.

Our contributions include: (1) The first large-scale investigation demonstrating that pure facial dynamics, completely separated from appearance, contain identity signals; (2) Evidence that Conformer's hybrid architecture optimally captures multi-scale facial patterns, outperforming both pure attention and convolution approaches; (3) A drift-to-noise ratio (DNR) metric that quantifies disentanglement quality and enables practical performance improvements; (4) Comprehensive analysis of temporal context and data requirements for robust dynamics-based identification.

\section{Related Work}

\subsection{Facial Dynamics for Identity Recognition}
Early work by~\cite{dobs2016identity} demonstrated that identity information content depends on facial movement type. Recent approaches have explored various representations: \cite{papadopoulos2022facegcn} used graph convolutional networks on facial keypoints, while~\cite{kay2023person} applied deep learning to micro-expressions to recognize people. However, these methods inherently mix static appearance with dynamic patterns, unable to determine whether recognition derives from motion or geometry.

\subsection{Disentangled Face Representations}
3D Morphable Models (3DMMs) offer a principled approach to separate facial components. The FLAME model~\cite{li2017flame} parameterizes faces into orthogonal subspaces: shape (identity-specific geometry), expression (dynamic deformations), and pose (rigid transformations). While FLAME provides the 3D model, we need a way to estimate its parameters from 2D video frames. For this, we use \textbf{VGGHeads} \cite{filntisis2022vggheads}, a deep neural network designed for head detection and 3D alignment.
VGGHeads integrates head detect/track and FLAME regression in one model, runs frame-wise at high throughput (critical for millions of frames across 1,429 speakers), and requires no multi-image identity constraints  \cite{sanyal2019learning} or scan supervision \cite{zielonka2022towards} or detail modeling and emotion consistency \cite{feng2021learning, danvevcek2022emoca}.

\subsection{Temporal Modeling for Facial Analysis}
Facial expressions exhibit complex multi-scale temporal dynamics: micro-expressions lasting 40-200ms reveal suppressed emotions \cite{chen2023smg, zhang2024review}, standard expressions spanning 0.5-4s convey intentional communication \cite{wang2024survey}, and conversational patterns extending beyond 4s encode interaction dynamics. Recent architectural advances have shown promise for capturing these hierarchical patterns. Conformer models~\cite{gulati2020conformer}, which elegantly combine self-attention mechanisms with convolutional layers, have achieved state-of-the-art results in speech recognition by effectively modeling both local and global dependencies. Their success extends to facial analysis, with applications in multimodal emotion classification~\cite{zhang2023uni2mul} and audio-driven animation~\cite{lin2024takin}.
\begin{figure*}[t]
  \centering
  \includegraphics[width=\textwidth]{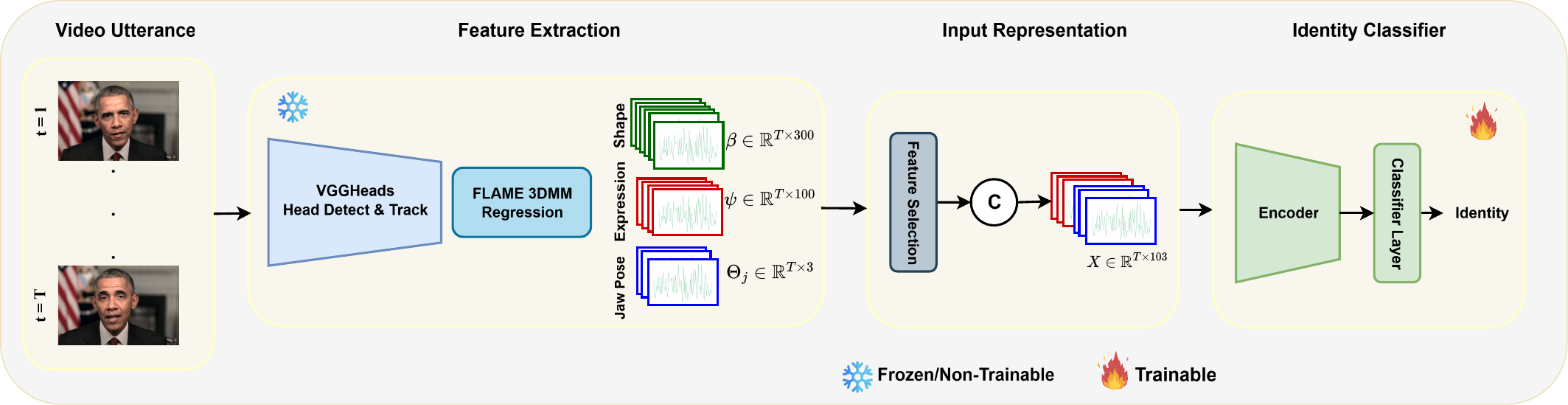}
  \caption{\textbf{Dynamics-only identification pipeline.} A video utterance is processed
  by a frozen front end (VGGHeads + FLAME) to obtain per-frame parameters. We retain only
  expression $\psi$ and jaw $\boldsymbol{\theta}_{j}$ to form $\mathbf{X}\in\mathbb{R}^{T\times 103}$, which a temporal encoder
  maps to an identity prediction.}

  \label{fig:diagram_wide}
\end{figure*}
\section{Methodology}

\subsection{Problem Formulation}
Given a video sequence $V = \{I_t\}_{t=1}^T$ of a person's face during conversation, our goal is to determine their identity $y \in \{1, ..., C\}$ using exclusively dynamic facial behavior, with complete removal of static facial structure. This requires three key components: (1) extracting facial representations that mathematically separate static shape from dynamic expression, (2) isolating purely dynamic features, and (3) learning person-specific temporal patterns from these dynamics alone. Our pipeline is shown in Figure ~\ref{fig:diagram_wide}.

\subsection{Dataset and Preprocessing}

We use a curated version of the CANDOR conversational corpus~\cite{reece2023candor} with 1429 speakers in natural, unscripted dyadic video calls. A \emph{session} denotes a unique online meeting; an \emph{utterance} is a short clip segmented by the speech transcript. The data is genuinely in-the-wild, with variation in lighting, background, occlusion, and head motion.This modified version of the CANDOR corpus exhibits a high degree of class imbalance.  The total number of video clips is 87,446 for training, 15,632 for validation, and 19,426 for testing. The number of utterances per speaker varies between 2 and 104, with an average of 62 and a standard deviation of 24.

To assess generalization, we evaluate on two complementary partitions:
\textbf{Group A (GA)}—678 speakers whose train/val/test utterances come from the \emph{same} session (intra-session identification), and
\textbf{Group B (GB)}—751 speakers with utterances drawn from \emph{different} sessions across splits (cross-session shift in appearance and environment).
This dual setup measures the upper bound under stable conditions (GA) and robustness to realistic session changes (GB).

\subsection{Disentangled Feature Extraction via FLAME}

\subsubsection{FLAME Representation}
FLAME~\cite{li2017flame} parameterizes a 3D head mesh through interpretable, low-dimensional components:
\begin{itemize}
    \item \textbf{Shape} ($\beta \in \mathbb{R}^{300}$): Static, person-specific geometry
    \item \textbf{Expression} ($\psi \in \mathbb{R}^{100}$): Dynamic, non-rigid deformations
    \item \textbf{Pose} ($\theta$): Articulated jaw rotation
\end{itemize}

The model generates a 3D mesh $M(\beta, \psi, \theta)$ where shape remains constant for an individual while expression and jaw vary with facial movement.

\subsubsection{Parameter Extraction Pipeline}
We employ VGGHeads~\cite{filntisis2022vggheads} for frame-wise FLAME parameter regression. Critically, we \textbf{discard} static shape $\beta$, global pose, and translation, retaining only dynamic components:

\begin{equation}
\mathbf{x}_t = [\psi_t, \theta_{j,t}] \in \mathbb{R}^{103}
\end{equation}

where $\psi_t$ represents 100 expression coefficients and $\theta_{j,t}$ represents 3 jaw rotation parameters at frame $t$. This yields a purely dynamic sequence $\mathbf{X} = \{\mathbf{x}_t\}_{t=1}^{T}$ for each utterance.
\subsection{Temporal Architecture Selection}
\subsubsection{Conformer Architecture}
We employ a compact Conformer~\cite{gulati2020conformer} that blends self–attention and convolution in a macaron-style arrangement, allowing the model to capture both long-range temporal dependencies and fine-grained local dynamics. In each block, attention (with positional cues) learns global relationships across the utterance, while a lightweight depthwise convolution emphasizes short, rapid movements typical of speech-driven facial motion. Residual connections and pre-normalization stabilize training. This hybrid design is well matched to our task: it preserves responsiveness to micro-scale articulations (jaw and lip dynamics) while tracking slower, conversational patterns that carry identity-specific style.
\subsubsection{Baseline Architectures}
For comprehensive evaluation, we implement multiple temporal encoders:
\begin{itemize}
    \item \textbf{GRU/MS-GRU}: Recurrent networks with single or multiple timescales
    \item \textbf{TCN/MS-TCN}: Temporal convolutions~\cite{lea2017temporal} with fixed or multiple kernel sizes $\{3,5,7,9\}$
    \item \textbf{Transformer}: Standard transformer encoder ~\cite{vaswani2017attention} with sinusoidal positional encoding
\end{itemize}
We use focal loss for training the baseline models which down-weights easy samples and concentrates optimization on misclassified, low-confidence examples, improving macro-F1 and cross-session recall.
\subsection{Two-Stage Training Framework}
\subsubsection{Stage 1: Representation Learning}
We train encoders using supervised contrastive learning~\cite{khosla2020supervised} to learn discriminative representations:
\begin{equation}
\mathcal{L}_{supcon} = \sum_{i \in I} \frac{-1}{|P(i)|} \sum_{p \in P(i)} \log \frac{\exp(z_i \cdot z_p / \tau)}{\sum_{a \in A(i)} \exp(z_i \cdot z_a / \tau)}
\end{equation}
where $P(i)$ denotes positive samples (same identity) for anchor $i$, $A(i)$ represents all samples except $i$, and $\tau = 0.07$ is the temperature parameter. A cross-batch memory queue maintains sufficient positive pairs even with limited samples per identity per batch.

\subsubsection{Stage 2: Classification}
We freeze the learned encoder and train a cosine-normalized linear classifier with cross-entropy loss and label smoothing ($\alpha = 0.1$) for improved calibration. This two-stage approach promotes learning of generalizable features before specializing for classification.
\subsection{Drift-to-Noise Ratio: Quantifying Disentanglement Quality}
While FLAME theoretically provides clean shape–expression separation, practical challenges arise in frame-wise estimation from unconstrained videos. A person's true facial shape should remain constant across sessions, but estimated parameters vary due to lighting changes, camera angles, and appearance variations (glasses, hairstyles), which can corrupt dynamic features. A necessary condition to compute DNR is having $\geq\!2$ distinct sessions per identity; therefore we estimate DNR only for Group~B (cross-session) in our data.

We quantify disentanglement quality with the Drift-to-Noise Ratio (DNR):
\begin{equation}
\text{DNR}(p) = \frac{\operatorname{median}_{\,s\neq s'\in S_p}\,\lVert \mu_{p,s}-\mu_{p,s'}\rVert_2}{\operatorname{mean}_{\,s\in S_p}\,\lVert \sigma_{p,s}\rVert_2 + \epsilon},
\end{equation}
where $S_p$ are sessions for person $p$, $\mu_{p,s}$ and $\sigma_{p,s}$ are the mean and standard deviation of estimated \emph{shape} parameters in session $s$, and $\epsilon=10^{-6}$ prevents division by zero. The numerator measures inter-session drift (ideally zero), the denominator the within-session noise baseline; DNR $\approx 1$ indicates acceptable disentanglement, whereas DNR $\gg 1$ signals shape leakage into dynamics (inspired by Fisher’s Discriminant Ratio (ratio of between-class scatter to within-class scatter) and Signal to Noise Ratio~\cite{fisher1936use,johnson2006signal}).

\section{Experiments}
\subsection{Protocol and Metrics}
We report overall and GA/GB accuracy and macro-F1. We train each model with AdamW (lr $1{\times}10^{-3}$, weight decay $1{\times}10^{-4}$), embedding dimension $128$, and batch size $1024$. Each utterance is a variable‐length sequence $\mathbf{X}=\{\mathbf{x}_t\}_{t=1}^{T}$ of FLAME dynamics.  
For identity classification we use a fixed maximum length of $L$. When $T\!<\!L$ we right–pad with zeros; when $T\!>\!L$ we take acenter crop of length $L$.%
\subsection{Architecture Performance Comparison}

Table~\ref{tab:main_results} presents comprehensive results across temporal architectures. The Conformer with supervised contrastive learning achieves best overall performance (60.09\% accuracy, 60.34\% macro F1), representing 458× improvement over chance (1/1429 = 0.07\%). This demonstrates that pure facial dynamics contain remarkably strong identity signals.Several key insights emerge from these results:

\begin{table*}[t]
\centering
\caption{Performance comparison across temporal architectures. Best results in each category are \textbf{bolded}.}
\label{tab:main_results}
\begin{tabular}{lccc|cc|cc}
\toprule
\multirow{2}{*}{Model} & \multirow{2}{*}{Loss} & \multicolumn{2}{c|}{\textbf{Overall}} & \multicolumn{2}{c|}{\textbf{Group A}} & \multicolumn{2}{c}{\textbf{Group B}} \\
& & Acc (\%) & F1 (\%) & Acc (\%) & F1 (\%) & Acc (\%) & F1 (\%) \\
\midrule
GRU & Focal & 45.64 & 46.95 & 80.36 & 58.33 & 35.20 & 19.93 \\
MS-GRU & Focal & 51.80 & 52.09 & 88.08 & 68.03 & 40.89 & 23.53 \\
TCN & Focal & 52.04 & 52.76 & 86.56 & 64.87 & 41.65 & 23.33 \\
MS-TCN(3,5,7) & Focal & 54.50 & 55.45 & 90.43 & 73.55 & 43.69 & 24.78 \\
MS-TCN(3,5,7,9) & Focal & 55.42 & 56.03 & 90.88 & 72.57 & 44.74 & 25.34 \\
Transformer & Focal & 55.14 & 55.80 & 89.48 & 71.06 & 44.81 & 25.26 \\
Conformer & Focal & 55.77 & 57.13 & 88.88 & 70.70 & 45.81 & 26.35 \\
\midrule
MS-TCN(3,5,7,9) & SupCon+Focal & 57.60 & 57.69 & 92.40 & 76.78 & 47.14 & 26.45 \\
Transformer & SupCon+Focal & 53.61 & 53.40 & 88.81 & 69.65 & 43.03 & 23.83 \\
\textbf{Conformer} & \textbf{SupCon+Focal} & \textbf{60.09} & \textbf{60.34} & \textbf{94.85} & \textbf{82.80} & \textbf{49.65} & \textbf{27.97} \\
\bottomrule
\end{tabular}
\end{table*}

\textbf{Conformer architecture excels:} The Conformer's hybrid design optimally balances local pattern detection (through convolution) with global context modeling (through attention). It outperforms MS-TCN by 2.49\% accuracy.

\textbf{Supervised contrastive learning provides selective benefits:} While Conformer gains substantially from contrastive pretraining (+4.32\% accuracy), Transformer performance unexpectedly degrades (-1.53\%). We hypothesize that pure self-attention, lacking convolution's inductive bias for local patterns, may overfit to spurious correlations in the contrastive embedding space. The Conformer's convolution module grounds abstract representations in concrete temporal dynamics.

\textbf{Group A achieves near-human performance:} The 94.85\% accuracy on Group A approaches human-level recognition from dynamics alone, suggesting our method successfully captures the full richness of individual expression patterns when environmental factors are controlled.

\subsection{Temporal Context Analysis}

Table~\ref{tab:seq_length} examines the relationship between sequence length and identification accuracy. Performance improves consistently with longer sequences, reaching 61.14\% at 900 frames (30 seconds at 30fps) suggests that extended temporal context captures additional identity-relevant patterns.

\begin{table}[t]
  \centering
  \caption{Effect of sequence length on Conformer with SupCon.}
  \label{tab:seq_length}
  \setlength{\tabcolsep}{1.9pt}
  \renewcommand{\arraystretch}{1.02}
  \begin{tabular}{lccccc c}
    \toprule
    \multirow{2}{*}{Length (frames)} & \multicolumn{2}{c}{Overall} & \multicolumn{2}{c}{GA} & \multicolumn{2}{c}{GB} \\
     & Acc (\%) & F1 (\%) & Acc (\%) & F1 (\%) & Acc (\%) & F1 (\%) \\
    \midrule
    300 (10s) & 60.29 & 60.51 & 94.47 & 80.60 & 50.01 & 28.02 \\
    480 (16s) & 60.09 & 60.34 & 94.85 & 76.78 & 49.65 & 27.97 \\
    600 (20s) & 60.81 & 61.17 & 94.67 & 81.45 & 50.62 & 28.40 \\
    900 (30s) & \textbf{61.14} & \textbf{61.29} & 94.66 & \textbf{83.86} & \textbf{50.98} & \textbf{28.63} \\
    \bottomrule
  \end{tabular}
\end{table}

\begin{figure*}[t]
  \centering
  \includegraphics[width=\textwidth]{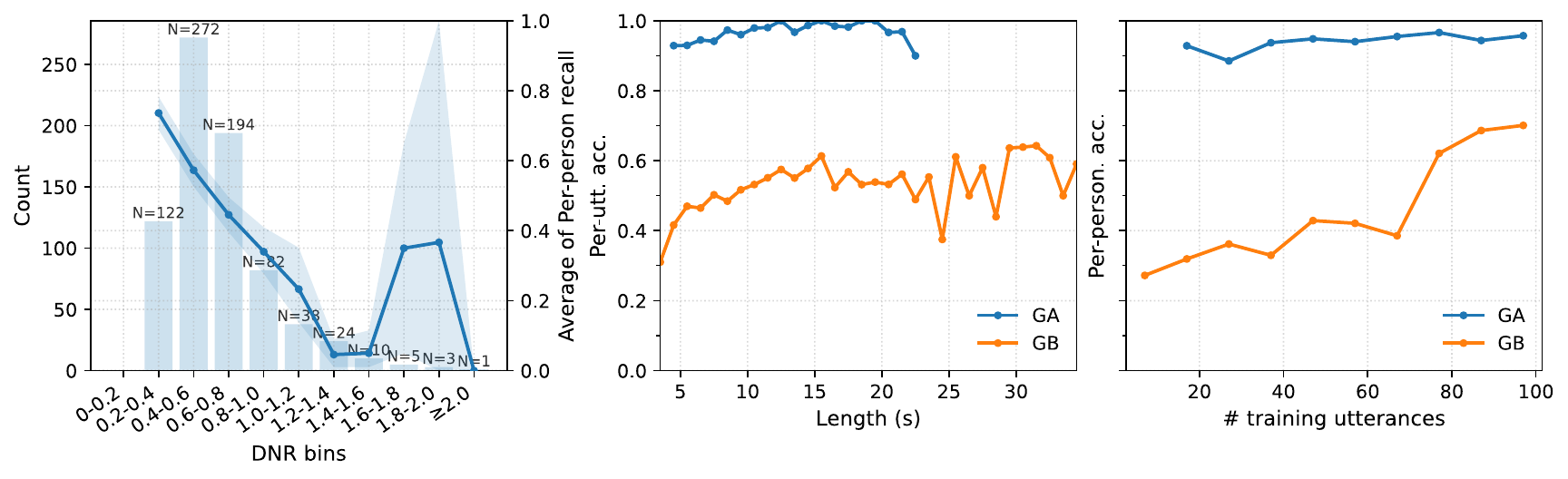}%
  \caption{\textbf{Performance predictors.}
  (left) \textbf{DNR vs.\ recall:} identities are binned by drift-to-noise ratio; line shows the mean per-person recall per bin, the shaded band is the 95\% CI, and bars denote the number of persons. Higher DNR corresponds to lower recall.
  (middle) \textbf{Accuracy vs.\ length:} per-utterance accuracy improves with longer clips; GA (same-session) remains above GB (cross-session) at all lengths.
  (right) \textbf{Accuracy vs.\ training utterances:} per-person accuracy increases with more training clips, with the largest gains for GB.}
  \label{fig:dnr_len_train}
\end{figure*}
\subsection{DNR Analysis and Performance Predictors}
We quantify how disentanglement reliability and the amount of temporal evidence predict recognition. For each identity we compute a \emph{drift-to-noise ratio} (DNR): the median inter-session distance between FLAME shape means, normalized by the average within-session variation. Identities are binned by DNR and we report binwise mean per-person recall with bootstrap 95\% CIs and counts. We also group test clips by duration to obtain micro-averaged per-utterance accuracy (GA vs.\ GB), and group identities by the number of training utterances to obtain per-person accuracy. Across all three analyses, higher DNR predicts lower recall, longer clips yield higher per-utterance accuracy (GA \(>\) GB at all lengths), and additional training utterances improve per-person accuracy—especially for GB.
\section{Discussion \& Conclusion}
This work provides the first large-scale evidence that dynamics-only facial signals—FLAME expression and jaw trajectories without appearance—encode distinctive person-specific information, revealing facial dynamics as underutilized behavioral signatures. Beyond their methodological interest, our results inform research on social perception, clinical assessment, and human–computer interaction, where appearance-agnostic cues are desirable. Although cross-session variation remains a deployment challenge, our analyses characterize performance as a function of temporal context and enrollment evidence (sequence length and number of training utterances), thereby offering quantitative guidance for practical operating points.

We further introduced the drift-to-noise ratio (DNR), which compares across-session changes in estimated shape to within-session variability. DNR exposes how external factors (e.g., lighting, viewpoint, accessories) perturb frame-wise 3D fitting and consequently affect the reliability of downstream dynamics. Empirically, a lower DNR correlates with a higher recognition accuracy when using dynamics-only features. Additionally, we excluded FLAME's eyeball parameters, though eye movements and gaze patterns exhibit person-specific characteristics that could provide complementary identity signals.

Future work will explore temporal consistency in other disentangled feature extraction models, self-supervised pretraining on larger datasets, eyeball parameters, and investigation of which specific dynamic patterns—smile dynamics, speech articulation, or emotional transitions—are most identity-discriminative. This work opens new avenues for understanding the behavioral signatures that make each person's expressions uniquely their own.

\bibliographystyle{IEEEtran}
\bibliography{refs}

\end{document}